\documentclass{article}

\usepackage{arxiv}

\usepackage[utf8]{inputenc} 
\usepackage[T1]{fontenc}    
\usepackage{hyperref}       
\usepackage{url}            
\usepackage{booktabs}       
\usepackage{amsfonts}       
\usepackage{nicefrac}       
\usepackage{microtype}      
\usepackage{lipsum}         
\usepackage{graphicx}
\usepackage{doi}
\usepackage{xspace}
\usepackage{amsmath}
\usepackage{amssymb}
\usepackage{amsthm}
\usepackage{bbm}
\usepackage{cleveref}       
\usepackage{bbm}
\usepackage{tabularx}
\usepackage{ltablex}
\usepackage{booktabs}
\usepackage{xspace}
\usepackage{fancyvrb}

\newcommand{\vbsd}{\ensuremath{\text{VBS}_{\text{dyn}}}\xspace}
\newcommand{\vbss}{\ensuremath{\text{VBS}_{\text{static}}}\xspace}

\title{Switching between Numerical Black-box Optimization Algorithms with Warm-starting Policies}

\date{}

\author{ 
    \href{https://orcid.org/0000-0003-2310-2862}{\includegraphics[scale=0.06]{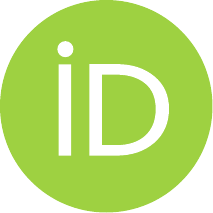}
    \hspace{1mm}Dominik Schr{\"o}der} \\
	LIACS, Leiden University\\
	Leiden, The Netherlands\\
	\texttt{dominik.schroeder94@gmail.com} \\
	\And
	\href{https://orcid.org/0000-0003-3040-7162}{\includegraphics[scale=0.06]{orcid.pdf}
    \hspace{1mm}Diederick Vermetten} \\
	LIACS, Leiden University\\
	Leiden, The Netherlands\\
	\texttt{d.l.vermetten@liacs.leidenuniv.nl} \\
	\And
	\href{https://orcid.org/0000-0002-4933-5181}{\includegraphics[scale=0.06]{orcid.pdf}
    \hspace{1mm}Hao Wang} \\
	LIACS, Leiden University\\
	Leiden, The Netherlands\\
	\texttt{h.wang@liacs.leidenuniv.nl} \\
	\And
	\href{https://orcid.org/0000-0002-4981-3227}{\includegraphics[scale=0.06]{orcid.pdf}
    \hspace{1mm}Carola Doerr} \\
    Sorbonne Universit\'e, CNRS, LIP6 \\
    Paris, France \\
	\texttt{Carola.Doerr@lip6.fr} \\
	\And
	\href{https://orcid.org/0000-0001-6768-1478}{\includegraphics[scale=0.06]{orcid.pdf}
    \hspace{1mm}Thomas B\"ack} \\
	LIACS, Leiden University\\
	Leiden, The Netherlands\\
	\texttt{t.h.w.baeck@liacs.leidenuniv.nl} \\
}

\renewcommand{\shorttitle}{Switching between Numerical Black-box Optimization Algorithms}

\begin{document}
\maketitle

\begin{abstract}
When solving optimization problems with black-box approaches, the algorithms gather valuable information about the problem instance during the optimization process. This information is used to adjust the distributions from which new solution candidates are sampled. In fact, a key objective in evolutionary computation is to identify the most effective ways to collect and exploit instance knowledge. However, while considerable work is devoted to adjusting hyper-parameters of black-box optimization algorithms on the fly or exchanging some of its modular components, we barely know how to effectively switch between different black-box optimization algorithms.

In this work, we build on the recent study of Vermetten et al. [GECCO 2020], who presented a data-driven approach to investigate promising switches between pairs of algorithms for numerical black-box optimization. We replicate their approach with a portfolio of five algorithms and investigate whether the predicted performance gains are realized when executing the most promising switches. Our results suggest that with a single switch between two algorithms, we outperform the best static choice among the five algorithms on 48 out of the 120 considered problem instances, the 24 BBOB functions in five different dimensions. We also show that for switching between BFGS and CMA-ES, a proper warm-starting of the parameters is crucial to realize high-performance gains. Lastly, with a sensitivity analysis, we find the actual performance gain per run is largely affected by the switching point, and in some cases, the switching point yielding the best actual performance differs from the one computed from the theoretical gain.
\end{abstract}

\section{Introduction}
Black-box optimization approaches are routinely applied in numerous applications every day. These applications span a great variety of industrial branches and academic domains. It is, therefore, not surprising that a plethora of algorithms exists, often developed with a specific application in mind.

Black-box optimization algorithms typically work in an iterative fashion. They decide for which solution candidates they wish to know the quality, evaluate these, and then use this information to adjust the strategy to decide upon the next search points. The choice of the solution candidates is often based on a stochastic procedure, effectively sampling the points from some random variables (which may or may not be explicitly modeled). Black-box optimization algorithms are hence inherently adaptive and adjust their sampling strategy during the optimization process. This adaptive behavior can be limited to controlling individual features of an otherwise fixed sampling distribution such as the search radius (``step size'' or ``mutation rate'' in evolutionary computation)~\cite{back1992self} or the covariance matrix in the case of the CMA-ES~\cite{DBLP:journals/ec/HansenO01}; these approaches are often referred to as \emph{parameter control}~\cite{EibenHM99,KarafotiasHE15,DoerrD18chapter}. But there are also attempts to dynamically replace whole modules of the algorithms, e.g., by replacing a greedy selection scheme with a non-greedy one or by switching on or off the use of crossover. Such strategies are studied under the umbrella of \emph{hyper-heuristics}~\cite{BurkeGHKOOQ13} or \emph{adaptive operator selection}~\cite{dacosta2008adaptive,sallam2018improved}, or they are implemented as part of a modular algorithm framework~\cite{DBLP:conf/gecco/VermettenRBD19}. More recently, attempts to actively learn dynamic algorithm designs via dedicated training steps are studied under the term \emph{dynamic algorithm configuration}~\cite{BiedenkappBEHL20,SharmaKLK19}.

In this work, we extend the scope of the dynamic choices by considering switching not only between different algorithm parameters or components but even between algorithms of very different types. To this end, we build on recent work~\cite{VermettenWBD20}, which suggests the great potential of such approaches even in the case that only a single switch between two algorithms is performed. This study by Vermetten et al. is based on the publicly available performance data repository of the BBOB function set (see Section~\ref{sec:BBOB} for details). Our main objective in this work is to investigate if these hypothetical performance gains identified in~\cite{VermettenWBD20} can indeed be realized in practice. To this end, we implement the switching strategies suggested by their approach and compare the achieved performance to that of the best static solver as well as to the ``theoretical'' performance computed by simply gluing the fixed-target performance curves of the chosen algorithms at a suitable switching point. Since these experiments require access to the algorithm implementations (which are unfortunately not available for most of the BBOB data files), we base our study on a portfolio of five quite diverse solvers: CMA-ES, BFGS, DE, PSO, and MLSL; see Section~\ref{sec:portfolio} for details and references.

For the design of the switching strategies, we face three main questions: `\textit{What algorithms to switch between?}', `\textit{When to switch?}',  and `\textit{How to perform the switch?}'. While we base the choice of the algorithms and the switching point on the approach suggested in~\cite{VermettenWBD20}, we experiment with different approaches to ``warm-start'' the second algorithm with information obtained from the partial optimization process of the first. Considering a switch between BFGS and CMA-ES, we show that it does not suffice to only initialize the search around the best-so-far solution(s) or to initialize (in addition) the step size, but that in order to realize the predicted performance gains, one also needs to warm-start the covariance matrix of the CMA-ES and the Hessian of BFGS, respectively.

All in all, we obtain encouraging results. In 48 out of the 120 considered examples (24 BBOB functions in 5 dimensions), we obtain a performance gain over the best static algorithm from our portfolio, and this is with a single switch. However, we also show that the performance of the individual runs can vary substantially and that the choice of the switching point using the approach from~\cite{VermettenWBD20} may not be optimal. At the time of writing, these observations have already led to follow-up works considering a \emph{per-run} switching strategy~\cite{DBLP:conf/ppsn/KostovskaJVNWED22}, as opposed to the \emph{per-instance} approach implemented in this present work.

\textbf{Related Work:}  Switching between different algorithm families is not a new idea. In particular, strategies to switch between population-based, more exploratory algorithms to  single-solution, exploitative local search methods have been investigated in~\cite{kelner2008hybrid, neri2012memetic}. Other successful examples of switching between different algorithm classes are available in the NGopt algorithm selection wizard~\cite{meunier2021black}. Unfortunately, apart from the code that is available open-access on GitHub~\cite{nevergrad}, the specifics or motivation for the implemented ``chaining'' procedures, as they are referred to in the Nevergrad environment, are not documented. 

For discrete black-box optimization scenarios, an approach similar to ours has been investigated in~\cite{ye2021leveraging}. The motivation of that study is similar to ours: we assume to be in a situation where some performance data is available, e.g., from previous experiments or from some benchmarking studies. We then need to decide which algorithm(s) to execute for a given problem instance. We can decide to switch between algorithms, but we cannot perform additional experiments as needed, for example, in dynamic algorithm configuration. Similar to our study, Ye et al. obtained promising performance of the switching strategies for some problems but also observed a big discrepancy between predicted and actual performance for others.   

\textbf{Availability of Code and Data:} 
Implementing suggestions made in~\cite{lopez2021reproducibility}, the full setup for reproducing all results presented in this paper can be found on zenodo~\cite{zenodo_dominik_dynAS}. This includes the code for the algorithm portfolio as well as the raw performance data and the scripts for processing and visualizing it.
The data of the larger static algorithm portfolio (described in Section~\ref{sec:portfolio}) is also available on the public repository of IOHanalyzer~\cite{wang2020iohanalyzer}, which can be found at \url{https://iohanalyzer.liacs.nl/} (the data set of each algorithm can be obtained by first selecting the ``IOH'' repository and then choosing the ``bbob'' data source).

\section{Background}
\label{sec:rel_work}

\subsection{The BBOB Problem Set}
\label{sec:BBOB}
For our experiments, we take the widely used noiseless Black-Box Optimization Benchmarking (BBOB) problem set~\cite{hansen:inria-00362633}, comprising 24 real-valued test functions to be minimized within the default search domain $[-5,5]^d$. The functions are chosen to cover a broad spectrum of problem characteristics that are expected to occur frequently in the continuous domain, e.g., separability and multi-modality. 
The dimensionality of the search domain can be chosen freely by the user. In this paper, we set the dimensionality to $d \in \{2, 3, 5, 10, 20\}$.  For each BBOB problem, different instances can be generated via transformations in the domain and range of the test functions that preserve the overall problem structure. In this paper, we take the first five instances of each function for the experimentation, i.e., the instance number $i \in \{1, 2, 3, 4, 5\}$. As the optimal function value $f_{\text{opt}}$ is known for each BBOB function, we commonly consider the gap between the best-so-far function value and the optimal one, $f_{\text{opt}}$, for measuring the progress. That is, we are mostly interested in the \emph{target precision} $\phi = f_{\text{best-so-far}} - f_{\text{opt}}$.

The BBOB test suite has been used for benchmarking workshops at academic conferences since 2009. As of March 2022, 232 different optimization algorithms and the corresponding empirical performance data have been submitted to the BBOB platform.\footnote{The data sets are available on \url{https://numbbo.github.io/data-archive/bbob/} and through the IOHanalyzer webinterface at \url{https://iohanalyzer.liacs.nl/} (as 'bbob' under 'dataset source').} The availability of this data makes the BBOB platform particularly useful for research on algorithm switching as it not only provides us with a diverse set of algorithms but also allows a more detailed analysis of different algorithmic search behaviors based on a large amount of available data.

\subsection{Empirical Performance Measure} 

In this work, we take a \textit{fixed-target} perspective to assess the empirical performance, where we measure the number of function evaluations that are required to reach a certain target precision. The \textit{hitting time} $T(A,f,d,\phi)$ is defined by the number of function evaluations that are performed by algorithm $A$ to reach a target precision $\phi$, where $f$ denotes the objective function, and $d$ denotes the dimensionality of the problem domain. If the algorithm did not find the target within the allocated budget, the hitting time is set to $T(A,f,d,\phi) = \infty$. Note that all algorithms considered in this study are stochastic so that the hitting time $T(A,f,d,\phi)$ is a random variable.

Since the optimization algorithms may not always reach the defined target precision, previous work on continuous black-box optimization commonly refers to the \textit{Expected Running Time} (ERT)~\cite{HansenABT22, Kerschke_2019}. Assuming the algorithm restarts for unsuccessful runs, the ERT estimates the expected hitting time of this restarting strategy by taking the ratio between the total number of function evaluations taken in all runs and the number of successful runs:
\begin{equation}
 \operatorname{ERT}(A,f,d,B,\phi) = \frac{\sum_{i}\textrm{min }\{(T_i(A,f,d,\phi)), B\}} {\sum_{i}\mathbbm{1}(T_i(A,f,d,\phi)<\infty)},  
\end{equation}
where $i$ denotes the $i$-th run of the algorithm on the problem (across all instances), $B$ denotes the maximal budget for function evaluations, and $\mathbbm{1}$ stands for the characteristic function.

\subsection{Theoretical Performance Gains}\label{sec:potential}
\begin{figure}
    \centering
    \includegraphics[width=\textwidth, trim=2mm 10mm 13mm 5mm, clip]{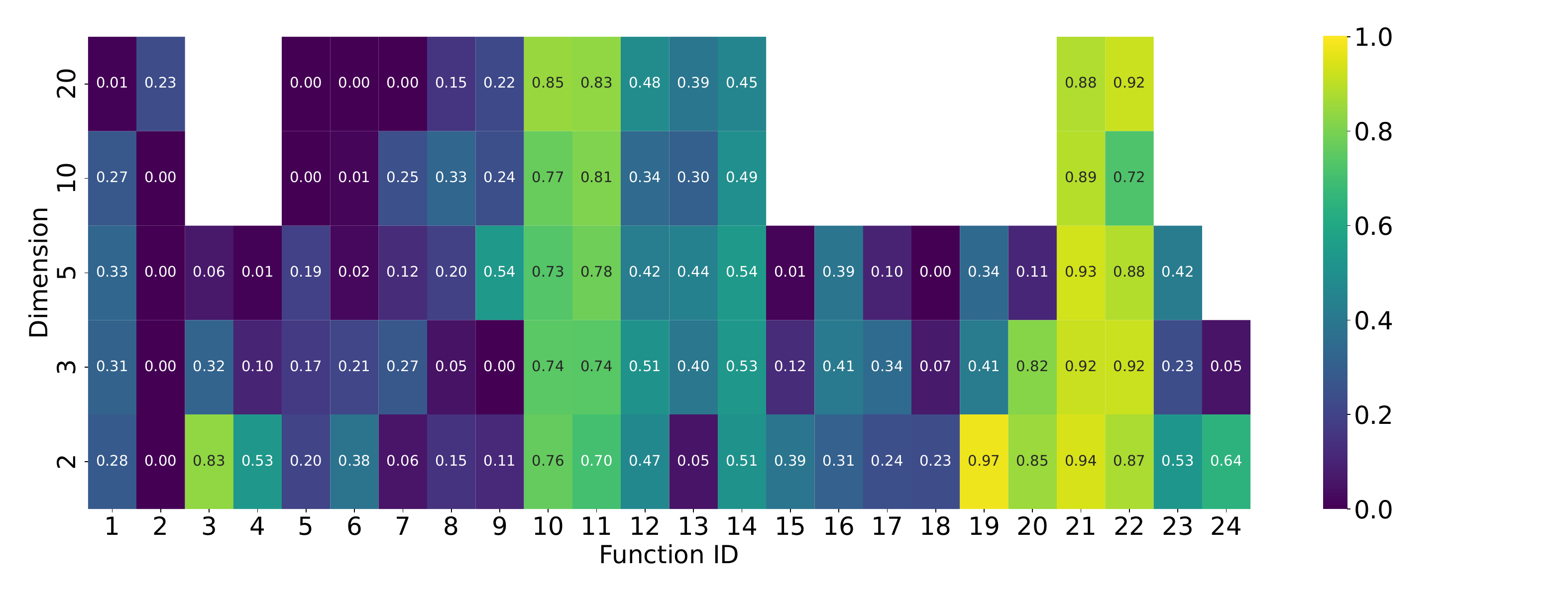}
    \caption{The theoretical performance gain of \vbsd over \vbss for the portfolio as described in  Sec.~\ref{sec:portfolio}, i.e., ($ERT(\vbss) - ERT_t(\vbsd)) / ERT(\vbss)$, in which the  best dynamic algorithm is the ($A_1, A_2$)-pair that maximized the theoretical gain for each combination of dimension and function. We compute these gain values on the BBOB test functions in several different dimensions. Please see Sec.~\ref{benchmarking-setup} for the detail regarding the generation procedure for the data set used here. 
    }
    \label{fig:Matrix-plot-potential-improvements}
\end{figure}

Previous work on switching between optimization algorithms~\cite{VermettenWBD20} has identified significant potential for performance improvement using a data-driven technique applied to a large set of performance data. The data from more than 100 algorithms submitted to the above-mentioned BBOB data repository was analyzed by defining a so-called ``theoretical performance'', which is defined based on the performance of its component algorithms. Assume we start the optimization process with an algorithm $A_1$ and switch to a different algorithm $A_2$ when $A_1$ hits a pre-defined target precision $\tau$ (the switching point, which is larger than the final target value $\phi$). Also, we assume that after the switching, $A_2$ would have the same internal states as if we ran $A_2$ from the beginning to hit $\tau$\footnote{Note that we are implicitly assuming that the algorithms also reach the same parts of the search-space, which might not be the case in practice.}. Under such assumptions, the \textbf{theoretical performance} of switching from $A_1$ to $A_2$ w.r.t. the switching point $\tau$ and the final target $\phi$ is:
\begin{equation}
    ERT_t(f, d, A_{1}, A_{2}, \tau, \phi) = ERT(A_{1}, f, d, \tau) + ERT(A_{2}, f, d, \phi) - ERT(A_{2}, f, d, \tau),
    \label{eqn:single-split-performance}
\end{equation}
where $f$ is a test function supported on a $d$-dimensional manifold, and the $ERT_t$ indicates that it is a theoretical measure. By calculating this theoretical value for all combinations of $A_1$, $A_2$, and $\tau$, we can find, for each function-dimension pair, the theoretically best dynamic algorithm switching procedure, which we refer to as the \emph{dynamic virtual best solver} (\vbsd). In the following discussion, when the $(A_1, A_2)$-pair of a \vbsd is clear from the context, we shall denote the theoretical gain alternatively by $ERT_t(\vbsd)$. Furthermore, we can compare the $ERT_t$ value of \vbsd to the ERT of the \emph{static virtual best solver} \vbss, i.e., the best single algorithm for the corresponding function-dimension pair in terms of the ERT measure, resulting in the \textbf{theoretical performance gain} of the dynamic switching approach w.r.t. the best single algorithm, e.g., taking a relative measure $(ERT(\vbss) - ERT_t(\vbsd)) / ERT(\vbss)$. Such a performance gain is illustrated in Fig.~\ref{fig:Matrix-plot-potential-improvements} for the five selected optimization algorithms of our algorithm portfolio (see next section for details), showing that for most functions, there is a significant theoretical gain brought by the dynamic switching approach. The theoretical advantage of \vbsd varies across different functions and dimensions, with a diminishing trend when the dimensionality increases. 
This seems to be caused by the fact that very few algorithms manage to reach the final target precision ($10^{-8}$) in high dimensions, and in the cases where they do, it is usually the case that one algorithm completely dominates all the others in terms of the ERT measure. 

Given a benchmarking data set, we find the best switching point $\tau$ for a specific algorithm pair $(A_1,A_2)$ and specific function $f$ with dimensionality $d$ by performing a line search on $\tau$. Afterward, for each function-dimension pair, we determine the theoretical \vbsd as the ($A_1$, $A_2$) pair that produces the smallest theoretical performance defined in Eq.~\eqref{eqn:single-split-performance}. 
We then take this found $(A_1,A_2,\tau)$ combination and run it on the corresponding function. From this, we again calculate ERT, which we will refer to as the \emph{achieved performance} of the $\vbsd$ for that function-dimension pair.
Similarly, the \textbf{relative performance improvement} over \vbss is defined as:
\begin{equation}
   \left(1-\frac{\texttt{min}(ERT(\vbss), ERT(\vbsd)))}{\texttt{max}(ERT(\vbss), ERT(\vbsd)))}\right) (2\cdot \mathbbm{1}_{ERT(\vbss) >  ERT(\vbsd))} - 1)
   \label{eq:relperf}
\end{equation} 
This measure takes values in $[-1,1]$ and is equivalent to the performance gain definition when an improvement is present but negative when the performance of \vbsd is worse than the \vbss.

\section{Our Use-Case with Five Diverse Algorithms}

\subsection{The Algorithm Portfolio}
\label{sec:portfolio}
\begin{figure}[!htp]
    \centering
    \includegraphics[width=.95\textwidth, trim=2mm 6mm 5mm 5mm, clip]{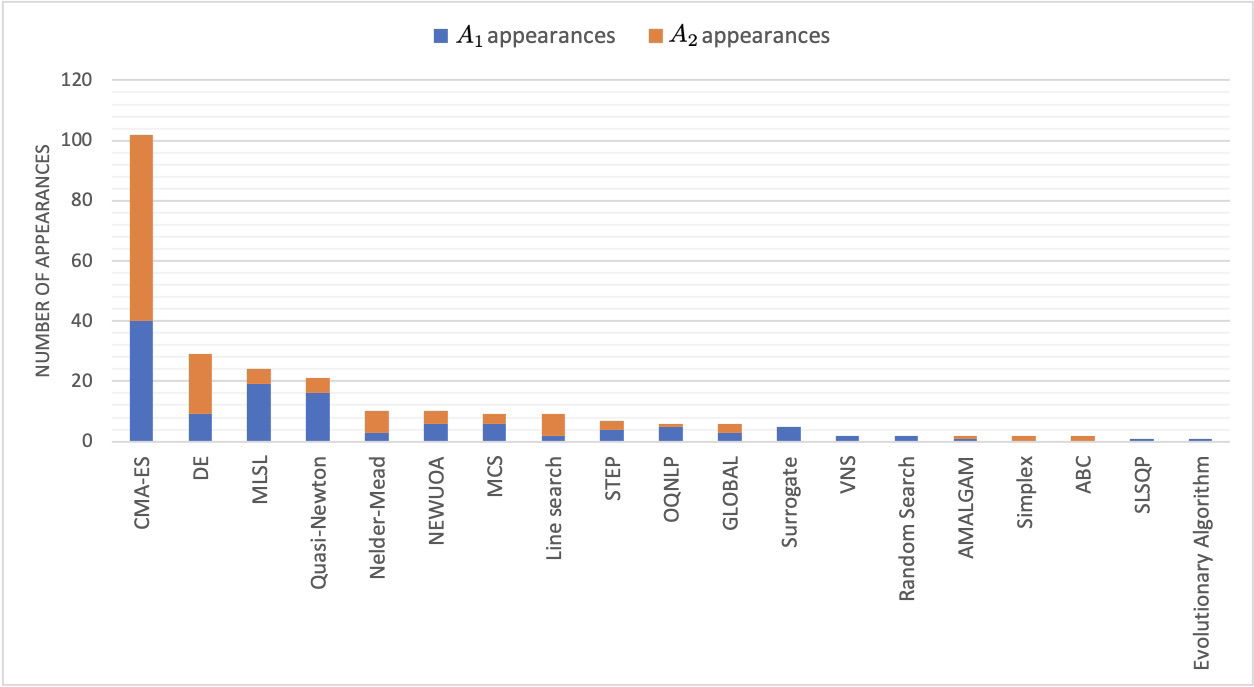}
    \caption{For constructing our algorithm portfolio, we examine the number of occurrences of each algorithm being the $A_1$ (in blue) or the $A_2$ (in orange) algorithm of the \vbsd on each BBOB test function for dimensions $d\in\{2, 3, 5, 10, 20\}$, as analyzed in~\cite{VermettenWBD20}.} 
    \label{fig:alg_family_a1_a2}
\end{figure}

We focus on a small algorithm portfolio rather than including the entirety of available solvers on the BBOB platform, which allows us to implement warm-starting mechanisms for each algorithm to verify whether the theoretical performance gains observed in previous work can be achieved in practice.
To ensure diversity within our portfolio, we include algorithms from different categories according to the taxonomy introduced in~\cite{DBLP:journals/corr/abs-1808-08818}, such as population-based and trajectory algorithms. Specifically, we select the algorithms that appear most frequently as either $A_1$ or $A_2$ of a \vbsd on some test function, based on the data set generated in~\cite{VermettenWBD20} (see Fig.~\ref{fig:alg_family_a1_a2}). This is based on the assumption that findings on the warm-starting procedure for the overall algorithm family will presumably be transferable to its variants as well, as long as they maintain similar internal parameters.
Concretely, our approach suggests the following five algorithms for our portfolio: 
\begin{itemize}
\item The Broyden-Fletcher-Goldfarb-Shanno (\textbf{BFGS}) algorithm~\cite{Broyden, Fletcher, Goldfarb, Shanno}, where the partial derivatives of the test function are approximated via the finite difference method. This algorithm represents the local search methods that occur often in Figure~\ref{fig:alg_family_a1_a2}.
\item The Multi-Level Single Linkage (\textbf{MLSL}) algorithm~\cite{DBLP:journals/mp/KanT87a}, which performs a series of local searches based on a clustering heuristic to avoid repeated search in the same area.
\item The Particle Swarm Optimization (\textbf{PSO}) \cite{DBLP:conf/icnn/KennedyE95} algorithm with a fully-connected neighbourhood topology (a.k.a.~global best) for the particles. While this algorithm doesn't appear in Figure~\ref{fig:alg_family_a1_a2}, it is a commonly used algorithm family, and thus we opted to include it in our portfolio.
\item The standard Covariance Matrix Adaption Evolution Strategy (\textbf{CMA-ES})~\cite{DBLP:journals/ec/HansenO01} without restarting heuristics.
\item A Differential Evolution (\textbf{DE})~\cite{DBLP:journals/jgo/StornP97} algorithm using binomial crossover and the mutation that uses the current best solution in the population (a.k.a.~DE/best/1). 
\end{itemize}

\subsection{Implementation Details}

We orientate ourselves towards existing algorithm implementations available from the BBOB data collection. In cases where the authors do not provide the code, we fall back on a different implementation while employing algorithm settings and parameters similar to the ones outlined in the respective submissions. 

For \textbf{BFGS}, we use the implementation from Scipy's \textit{optimize} module \cite{2020SciPy-NMeth}. Similar to \cite{DBLP:journals/corr/Baudis14}, we keep the algorithm's default settings. The major difference to their implementation is that we refrain from basin hopping as a restart strategy since, from the initial testing on BBOB, basin hopping often generates many numerical warnings. Hence we decided to take it out from the implementation of BFGS to ensure the reliability of the results.
After some initial experiments on the BBOB problems, we changed the gradient tolerance from $10^{-5}$ to $10^{-10}$ to prevent the BFGS algorithm from stopping too early. The initial search point is sampled randomly in~$[-5, 5]^d$. 

\begin{table}[t]
\footnotesize
\caption{The number of (function, dimension) pairs for which the ($A_1$, $A_2$)-pair is the \vbsd, counting only use cases for which the improvement over \vbss, the best static algorithm from the portfolio, is at least 1\%.
The final target precision is set to $\phi = 10^{-8}$. The most promising ($A_1$, $A_2$)-pairs 5 or more use-cases are marked in boldface.} 
\label{tab:combinations}
\centering
\begin{tabular}{|l|l|r|l|}
\hline
\textbf{$A_1$} & \textbf{$A_2$} & \textbf{\begin{tabular}[c]{@{}l@{}}Number of \\ use cases\end{tabular}} & \textbf{Functions and dimensions}                                       \\ \hline
CMA-ES                & DE                    & 1                                                                       &      F23 {[}5D{]}                                             \\
CMA-ES                & PSO                   & 0                                                                       & -                                                                       \\
CMA-ES                & MLSL                  & 1                                                                       & F6 {[}5D{]}                                                                       \\
\textbf{CMA-ES}                & \textbf{BFGS}                  & 5                                                                      & F8 {[}all dimensions{]}, F9 {[}20D{]}                          \\ \hline

\textbf{PSO}                   & \textbf{CMA-ES}                & 9                                                                       & F7, F16 {[}2D{]}, F3, F4, F15, F17, F18, F19 {[}3D{]}, F6 {[}10D{]}                 \\
\textbf{PSO}                   & \textbf{DE}                    & 7                                                                       & F42, F15, F17, F18, F22 2D, F16 3D, F3 F4 F19 5D                                 \\
PSO                   & MLSL                  & 0                                                                       & -                                                                       \\
PSO                   & BFGS                  & 3                                                                       & F1 {[}10D{]} F3, F5 2D                                                            \\ \hline
MLSL                  & CMA-ES                & 5                                                                       & F23 {[}2D, 3D{]}, F21 {[}10D{]}                            \\
MLSL                  & DE                    & 3                                                                       & F20, F21 {[}2D{]}                                                            \\
MLSL                  & PSO                   & 3                                                                       & F24 {[}2D{]}                                      \\
\textbf{MLSL}                  & \textbf{BFGS}                  & 9                                                                       & F19 2D, F20-23 3D, F5 F2 5D, F2, F21 20D\\ \hline
\textbf{DE}                    & \textbf{CMA-ES}                & 5                                                                       & F13 {[}2D{]}, F16 {[}5D{]}, F7  {[}10D{]}                                        \\
DE                    & PSO                   & 3                                                                       &  F15 {[}5D{]}, F7 {[}3D{]}                                      \\
DE                    & MLSL                  & 0                                                                       & -                                                                       \\
DE                    & BFGS                  & 5                                                                       & F1 {[}2D, 3D, 5D{]}, F9 2D, F5 {[}3D{]}                                                \\ \hline
\textbf{BFGS}                  & \textbf{CMA-ES}                & 23                                                                      & F5, F6, F10-F14 {[}various dimensions{]}                      \\
BFGS                  & DE                    & 1                                                                       & F22 10D                                                                       \\
BFGS                  & PSO                   & 0                                                                      & F20 5D                                                                     \\
BFGS                  & MLSL                  & 3                                                                       & F1 20D, F22 5D, F12 3D, F10-12 2D       \\ \hline
\end{tabular}
\end{table}

We use our own implementation of \textbf{MLSL}. The hyperparameters are set according to the BBOB submission \cite{DBLP:conf/gecco/Pal13}, with a population size of $50d$ and $10\%$ of the function evaluation budget being allocated to the local search.
The major difference in our Python implementation is that we use the \textit{Powell} method from Scipy's \textit{optimize} module for the local search routine rather than MATLAB's \verb|fmincon| interior-point method used in~\cite{DBLP:conf/gecco/Pal13} We chose Powell among the optimizers in the free Scipy optimize module because it suffered least from numerical issues in our (non-documented) test series on BBOB.
We use the default settings to run the Powell method, except for a reduced fitness tolerance parameter $f_\text{{tol}} = 10^{-8}$.

For \textbf{PSO}, we also use our own implementation as we cannot find the original code from the submission~\cite{DBLP:conf/gecco/El-AbdK09a}. In our implementation, we use the \textit{global best} neighborhood topology for the particles. The swarm size accounts for $40$ particles, which are sampled uniformly at random in $[-5, 5]^d$. We initialize the velocity randomly in $[-1, 1]^d$ to favor exploiting search behavior. The learning rate for updating the velocity is set to $1.4944$, as recommended in~\cite{BonyadiM17}.
In terms of boundary handling, particles that violate a constraint are clipped to the corresponding boundary, where their velocities are set to zero. Also, the velocities of particles are always clipped to $[-5, 5]$. The inertia weight $\omega$, which dampens the velocity update, is essential for preventing premature convergence, is adjusted using a linearly decreasing schedule: $\omega = 0.9 - 0.8t/B$, where $t$ is the number of function evaluations used so far and $B$ is the maximum budget. 

We take the modular CMA-ES framework~\cite{DBLP:conf/ssci/Rijn0LB16} for the implementation of the standard \textbf{CMA-ES} algorithm, where we use the default setting for the population size, i.e., $\lambda = 4 + \left \lfloor{3 \cdot \log d}\right \rfloor$. Also, we took the setting for hyperparameters as suggested in~\cite{Hansen06}. The modules for additional features, such as increasing population size and random restarts, are all turned off. The step size $\sigma$ is initially set to $0.2$ times the length of search space (assuming equal sized box-constraints), so for the BBOB functions, this results in $\sigma=2$. Lastly, the center of mass of the population is sampled uniformly at random in the whole search space.

As \textbf{DE} implementation, we employ the version that is available via Scipy's \textit{optimize} module~\cite{2020SciPy-NMeth}. Most parameters are left in their default settings: most importantly, we use the binomial crossover operator with a crossover rate of $0.7$ and the so-called ``DE/best/1'' mutation operator, which only uses the current best solution in the population for the mutation. We set the population size to $5d$ and reduce the convergence tolerance from $10^{-2}$ to $ 10^{-12}$ to prevent the DE algorithm from terminating too early before the depletion of the function evaluation budget. The mutation scaling factor is sampled uniformly from the interval $[0.5, 1]$ at each generation.

\subsection{Benchmarking Setup} \label{benchmarking-setup}
We run all five algorithms on all 24 functions within the BBOB test suite. We set the dimensionality to $d \in \{2, 3, 5, 10, 20\}$. Similar to a large part of existing research, the algorithms are run on each of the first five problem instances, $i \in \{1, 2, 3, 4, 5\}$. We perform 10 independent runs on each instance, resulting in a total of 50 runs per function-dimension pair. Runs are stopped when the final target precision $\phi = 10^{-8}$ is hit or when the budget for function evaluations $B = 10\,000d$ is depleted.

\subsection{Warm-starting Procedures}
\label{sec:warmstart}

While it is typically straightforward to initialize the second algorithm around the best point(s) found by its predecessor, it can be substantially more challenging to configure its free parameters, such as, in the case of the CMA-ES, the step-size or the covariance matrix. In general, when the algorithm contains some type of adaptive control mechanism, one would ideally use the already evaluated samples to ``warm-start'' the parameters controlling the search behavior. How to best do this is not always obvious.

In our use-cases, we use the following strategies to warm-start the algorithms. These are essentially based on preliminary experiments and best guesses based on our experience with these search strategies. 
\begin{itemize}
    \item CMA-ES: The center of mass is placed at the best-so-far point found by the original algorithm. The initial step size is determined based on the three best points found in the run of the previous algorithm (more details will be given in Section~\ref{subsec:CMA-BFGS}, where the modification for the particular case of switching to CMA-ES from BFGS is also discussed).
    \item BFGS: The starting point for the BFGS is set at the best-so-far point found by the original algorithm. Modification in the case of starting from CMA-ES is discussed in Section~\ref{subsec:CMA-BFGS}
    \item DE / PSO: we draw the starting population of these algorithms uniformly at random in the hyperbox $\left[\mathbf{x}^*-\mathbf{\eta}, \mathbf{x}^*+\mathbf{\eta}\right]$, where $\eta=0.1\cdot \mathbf{1}$ is chosen via preliminary experiments and $\mathbf{x}^*$ is the best point found by the previous algorithm and  $\mathbf{1}$ is the all-ones vector in dimension $d$.
    \item MLSL: If a population is present from the previous algorithm, this is set as the current population; we do not warm-start the algorithm otherwise. 
\end{itemize}

\section{Results}
\label{sec:results}

\subsection{Identifying promising $(A_1,A_2)$-pairs} 
\label{sec:data-driven-analysis}

Taking the data set generated by the procedure outlined in Section~\ref{sec:potential}, we compute the theoretical performance gain defined in Eq.~\eqref{eqn:single-split-performance} to determine the best possible switching combinations for each pair of test function and dimension. In Table~\ref{tab:combinations} we summarize the number of (function, dimension)-pairs for which \vbsd is a combination of the two given algorithms $A_1$ and $A_2$, with an advantage of at least 1\% over the best static choice, \vbss.  
We observe that combinations of BFGS and CMA-ES constitute the majority of these cases (for 28 out of the 120 studied examples, the \vbsd is composed of these two algorithms). Other algorithm combinations achieve the best theoretical performance for another 58 cases. In particular, switching from PSO to CMA-ES seems promising, according to the theoretical performance gain, for 9 cases, and switching from MLSL to BFGS seems promising for another 9 (function, dimension) pairs. We have a total number of 86 use cases for which the theoretical performance gain of the \vbsd over the \vbss is at least 1\%.

\subsection{Actual Performance Gains}
\label{sec:actual}

\begin{figure}[!t]
    \centering
    \includegraphics[width=\textwidth]{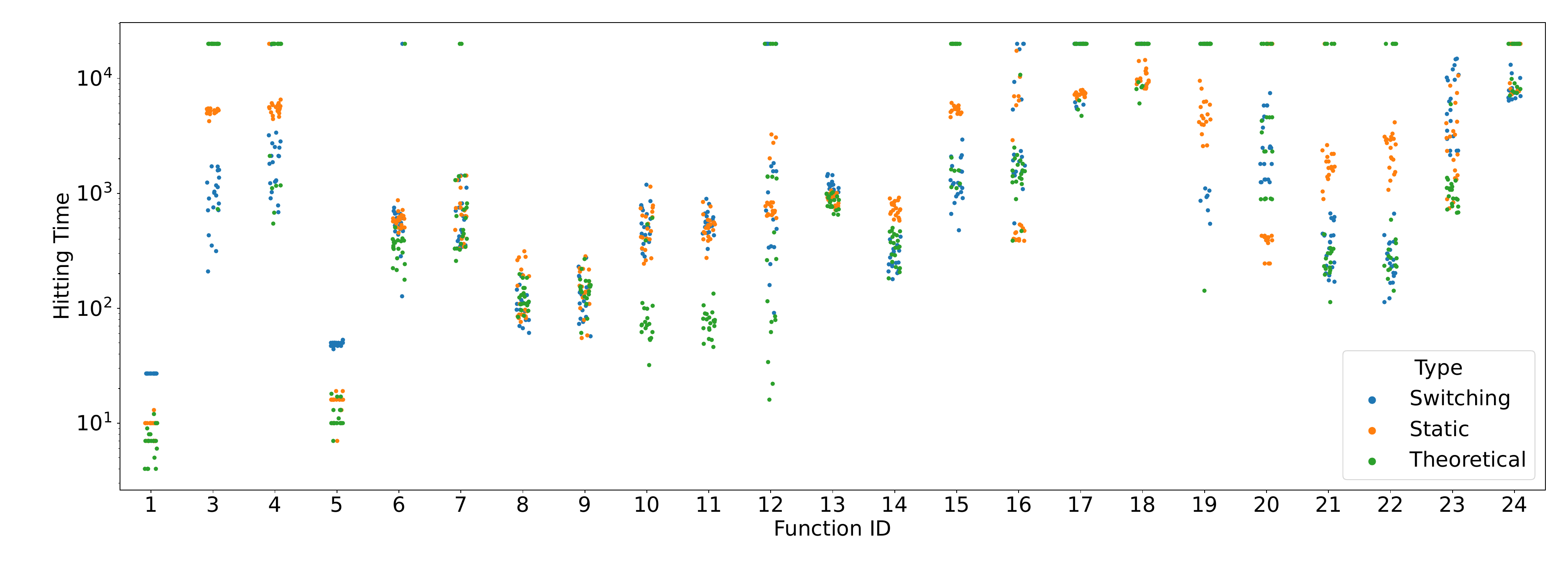}
    \caption{In 2D, for each BBOB test function (x-axis), the actual performance measured as hitting time of $1e^{-8}$ of the \vbsd (blue) is compared to its theoretical performance (green) and the \vbss (orange). All algorithms were run according to the settings outlined in Sec.~\ref{benchmarking-setup}. Each dot corresponds to 1 run, where for the theoretical values, each run from the first algorithm is combined with the corresponding run from the second algorithm.}
    \label{fig:ERT-comparison-d2}
\end{figure}

In Fig.~\ref{fig:ERT-comparison-d2} we show the actual performance gain calculated after executing the \vbsd (determined by the theoretical performance gain) on each function in 2D, and compare it with the theoretical performance prediction. We see several cases in which the actual performance $ERT(\vbsd)$ matches or even improves on the theoretical one $ERT_t(\vbsd)$. This is the case, e.g., for $F3$, $F6$-$F9$, $F14$, $F17$, and $F24$. However, there are also cases where the actual performance is worse than the theoretical one or even worse than the ERT value of the best static solver, the \vbss. This holds in particular for $F1$, $F4$, $F5$, $F15$, and $F20$. 

\begin{figure}[!h]
    \centering
    \includegraphics[width=.8\textwidth]{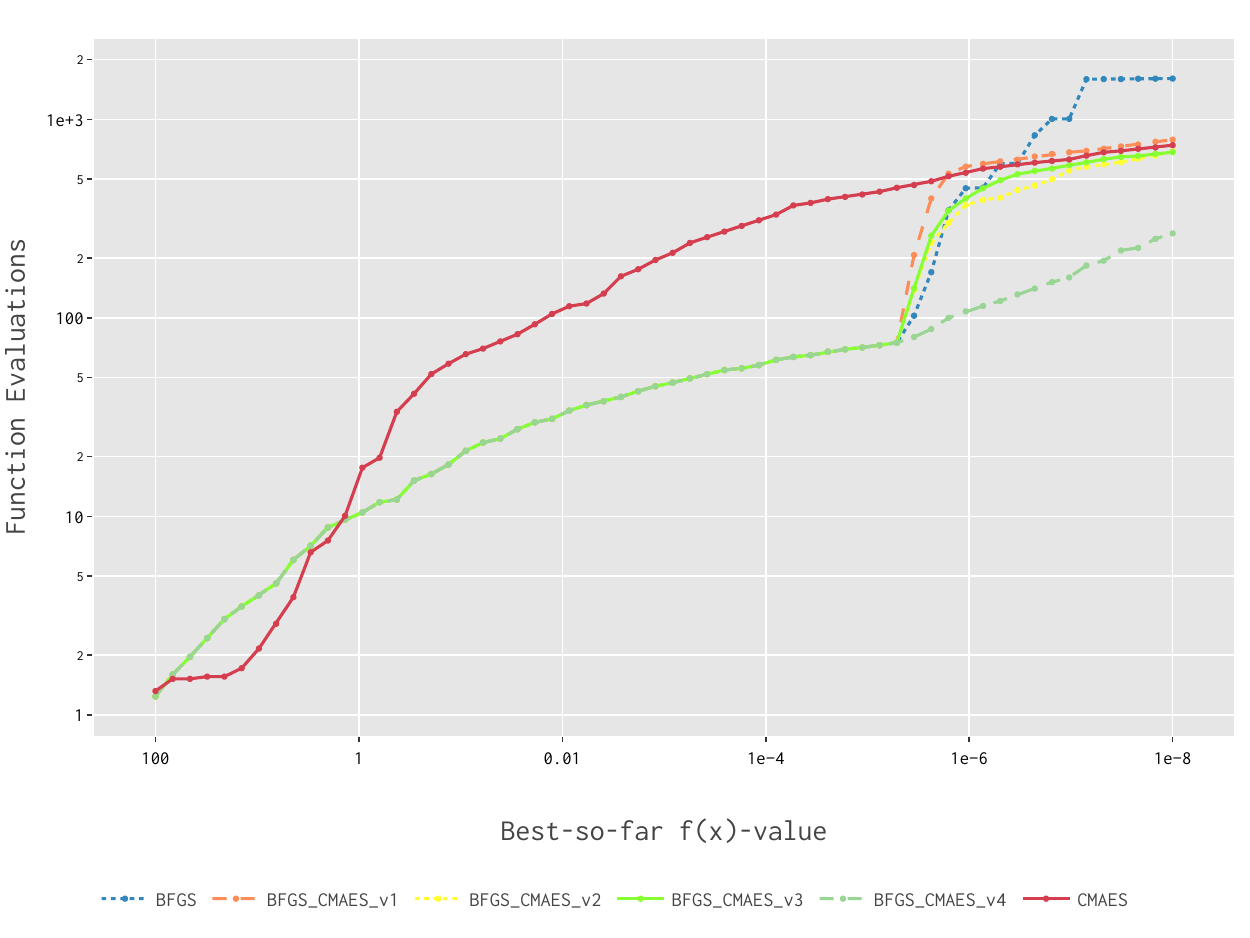}
    \caption{ERT chart for BFGS (blue), CMA-ES (red), and switching from BFGS to CMA-ES with different warm-starting routines on function $F14$ in 2D. The switching point is set to $\tau=10^{-5.4}$. Warm-starting routines all pass the best-so-far point of BFGS to the CMA-ES. V2 sets $\sigma=0.01$, while V3 and V4 set it based on the 3 best points from the BFGS trajectory. V4 also sets the covariance matrix to the inverse Hessian from the last step of BFGS. ERT values are computed from 25 independent runs. 
    }
    \label{fig:ERTs-bfgs-cmaes}
\end{figure}

For the function-dimension pairs where we observe a large gain over the \vbss, it is often the case that the ERT curve of $A_1$ is continued with that of $A_2$ after shifting the latter down to the switching point. This can be seen, for instance, when switching from BFGS to CMA-ES at $\tau=10^{-5.4}$ on function $F14$ in 2D; see Fig.~\ref{fig:ERTs-bfgs-cmaes}.

\begin{figure}
    \centering
    \includegraphics[width=\textwidth,  trim=2mm 10mm 11mm 5mm, clip]{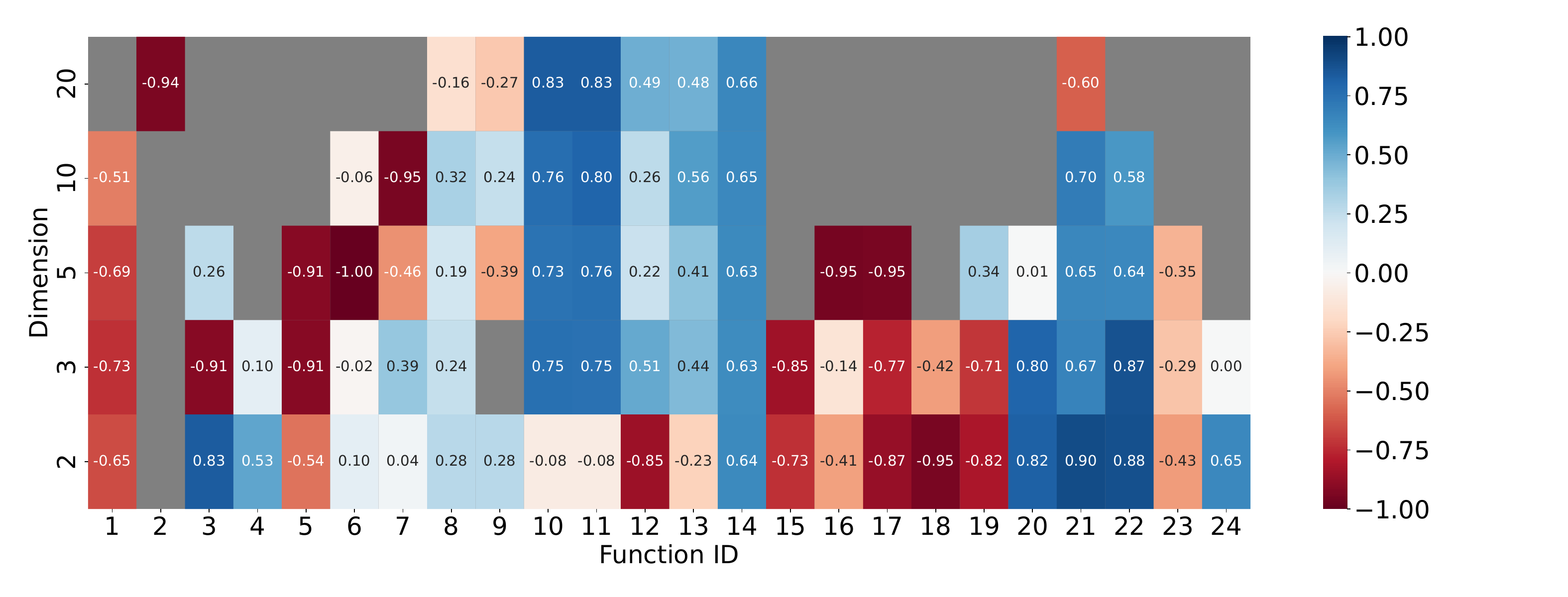}
    \caption{The actual performance gain of the \vbsd over the \vbss within our portfolio for all functions and dimensions, calculated according to Equation~\ref{eq:relperf}. Grey tiles indicate (function, dimension) pairs for which the theoretical performance prediction did not show an advantage of the \vbsd over the \vbss. Negative values correspond to cases for which the actual performance of the \vbsd is worse than the \vbss, despite a positive theoretical performance gain.
    \label{fig:ERT-impro-heatmap}
    }
\end{figure}

Summarizing over all the use cases, we found that the \vbsd outperforms the \vbss on 48 out of the 86 identified use cases (see Fig.~\ref{fig:ERT-impro-heatmap}). The highest actual performance gain (90\% relative reduction of the ERT w.r.t. the \vbss) is observed when switching from PSO to DE on $F21$ in 2D. Also, switching from BFGS to CMA-ES on functions $F10$-$F14$ produces a high actual performance gain while the reverse switching scheme, i.e., from CMA-ES to BFGS, still achieves significant gain on functions $F8$ and $F9$.
A key observation from our experiments is that, as might be expected, the actual performance gain can only be realized when $A_2$ is properly warm-started. We will investigate the impact of the warm-starting procedure in the following sections. It should be noted, however, that there are some use-cases where the predicted performance can not at all be met when switching between the selected algorithms, indicating that we may either need different warm-staring techniques or that some other assumptions made when gluing together the performance curves in the theoretical performance prediction are not met. A possible explanation could also be that the algorithms focus on different parts of the search space. We leave this question for future investigation, as it would require a thorough investigation that -- in light of the high stochasticity of the algorithms~\cite{VermettenW0DB22} -- should be based on many more runs than what our data offer.

\subsection{Zooming into Switching between CMA-ES and BFGS} 
\label{subsec:CMA-BFGS}

\begin{figure}[!t]
    \centering
    \includegraphics[width=0.98\textwidth]{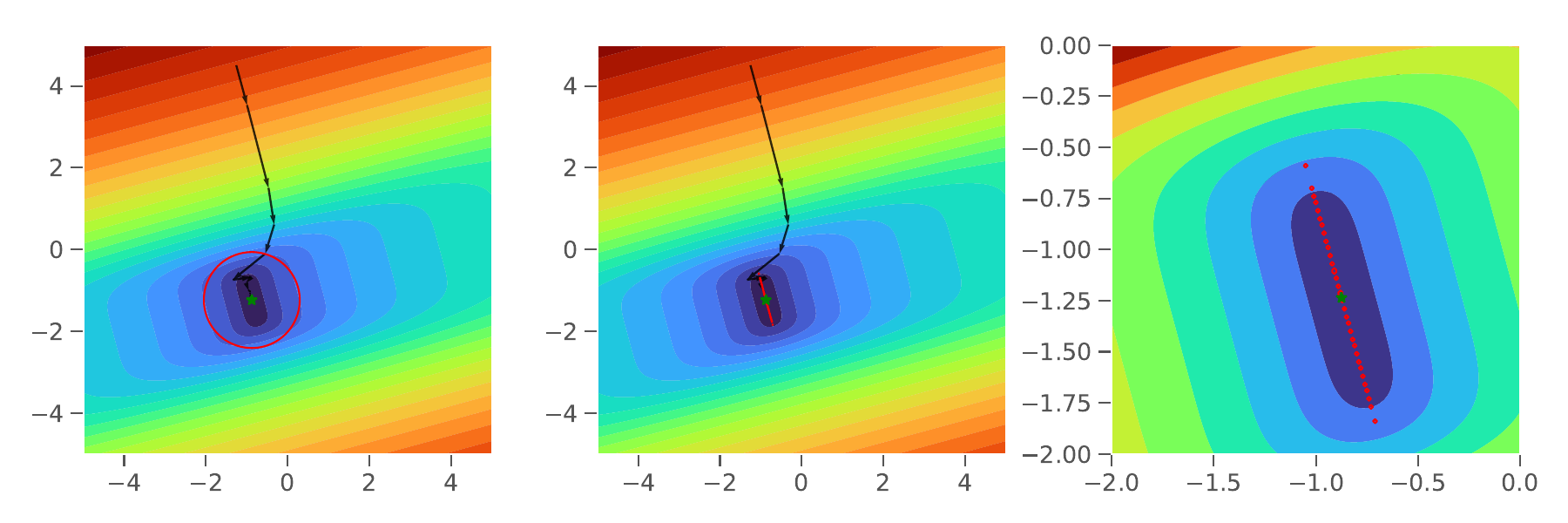}
    \caption{Contour plot of the covariance matrix of CMA-ES immediately after warm-starting from a BFGS-run (F14 in 2D, instance 1, with $\tau=10^{-5.4}$). The black line indicates the trajectory followed by BFGS, and the green star indicates the best-so-far point found when the switch occurs. The red lines indicate areas of equal probability for the CMA-ES to generate offspring. The leftmost figure shows the case where covariance and step size are not warm-started. The two right images show the result when incorporating the warm-starting. The right-most figure is a zoomed-in version of the middle one.}
    \label{fig:countour_cma_basic_new}
\end{figure}

To illustrate the impact of the warm-starting procedure on the optimization process, we can zoom in on the cases where we switch between CMA-ES and BFGS. For this use case, a first, na\"ive warm-starting procedure would simply select the best point in the CMA-ES population as the starting point for BFGS, and similarly the current point in BFGS as the center for the new CMA-ES population in the reverse case, i.e., when switching from BFGS to CMA-ES. As expected, considering only the search point ignores a large fraction of the information learned during the search and results in poor performance of the chained algorithms. In Fig.~\ref{fig:countour_cma_basic_new} (the leftmost sub-plot), we show the case where CMA-ES is initialized from only the best-so-far point of BFGS, so the direction of the search and the corresponding step-sizes are not taken into account.

To initialize the covariance matrix, we can use the fact that for convex-quadratic functions, the covariance matrix directly relates to the inverse Hessian of the objective function, which is approximated during the BFGS procedure~\cite{DBLP:conf/ppsn/GlasmachersK20, DBLP:journals/corr/Hansen16a, DBLP:journals/tcs/ShirY20}. 
While not all BBOB test functions are convex-quadratic, it is likely that the local landscape at the time of algorithm switching resembles a quadratic shape when the step size is sufficiently small. As such, we will use the approximated inverse Hessian matrix of BFGS for warm-staring the covariance matrix of CMA-ES, and we do the reverse (with some scaling factor $\beta$) when switching from CMA-ES to BFGS.

\begin{table}
\scriptsize
\renewcommand{\baselinestretch}{1.0}
\caption{ERT values for switching between BFGS and CMA-ES on selected function-dimension pairs. The target precision has been set to $\phi = 10^{-8}$. Performance gain values are calculated using Equation~\ref{eq:relperf} and Equation~\ref{eqn:single-split-performance} respectively for the last two columns. We mark the cases in boldface where the actual performance $ERT(\vbsd)$ is even better (lower) than the theoretical one $ERT_t(\vbsd)$.}
\begin{center}
\label{tab:bfgs_cmaes}
\begin{tabular}{|l|l|c|c|c|r|r|r|r|r|}
\hline
\textbf{$A_1$} & \textbf{$A_2$} & F & d & {$\log_{10}(\mathbf{\tau})$} & \multicolumn{1}{l|}{\begin{tabular}[c]{@{}l@{}}ERT(\vbss)\end{tabular}} & \multicolumn{1}{l|}{\begin{tabular}[c]{@{}l@{}} $ERT_t(\vbsd)$\end{tabular}} & \multicolumn{1}{l|}{\begin{tabular}[c]{@{}l@{}} $ERT(\vbsd)$\end{tabular}} & \begin{tabular}[c]{@{}l@{}} performance \\gain \end{tabular} & \begin{tabular}[c]{@{}l@{}}Theoretical \\ performance \\gain\end{tabular} \\ \hline
BFGS & CMA & 6 & 3 & -4.40 & 978.86 & 776.08 & 998.78 & -0.02 & 0.21 \\
BFGS & CMA & 8 & 20 & 0.80 & 42 976.24 & 36 362.14 & 51 146.10 & -0.16 & 0.15 \\
\textbf{BFGS} & \textbf{CMA} & \textbf{10} & \textbf{3} & -4.00 & 1 103.88 & 283.78 & 275.04 & 0.75 & 0.74 \\
BFGS & CMA & 10 & 5 & -3.20 & 2 168.38 & 578.44 & 575.80 & 0.73 & 0.73 \\
BFGS & CMA & 10 & 10 & -3.20 & 6 065.80 & 1 419.86 & 1 453.20 & 0.76 & 0.77 \\
BFGS & CMA & 10 & 20 & -4.40 & 19 362.58 & 2 971.24 & 3 284.84 & 0.83 & 0.85 \\
\textbf{BFGS} & \textbf{CMA} & \textbf{11} & \textbf{3} & -3.20 & 1 185.78 & 307.88 & 299.36 & 0.75 & 0.74 \\
BFGS & CMA & 11 & 5 & -3.20 & 2 348.60 & 514.14 & 574.72 & 0.76 & 0.78 \\
BFGS & CMA & 11 & 10 & -3.40 & 5 795.16 & 1 112.28 & 1 144.90 & 0.80 & 0.81 \\
BFGS & CMA & 11 & 20 & -3.40 & 15 572.26 & 2 607.96 & 2 611.94 & 0.83 & 0.83 \\
BFGS & CMA & 12 & 3 & -4.40 & 4 247.06 & 2 075.12 & 2 081.70 & 0.51 & 0.51 \\
BFGS & CMA & 12 & 5 & -4.40 & 4 284.50 & 2 484.00 & 3 322.92 & 0.22 & 0.42 \\
BFGS & CMA & 12 & 10 & -2.80 & 12 545.96 & 8 260.32 & 9 251.04 & 0.26 & 0.34 \\
\textbf{BFGS} & \textbf{CMA} & \textbf{12} & \textbf{20} & -3.80 & 30 700.28 & 15 856.96 & 15 801.50 & 0.49 & 0.48 \\
\textbf{BFGS} & \textbf{CMA} & \textbf{13} & \textbf{3} & -2.40 & 1 768.28 & 1 064.96 & 987.64 & 0.44 & 0.40 \\
BFGS & CMA & 13 & 5 & -2.60 & 3 635.46 & 2 049.82 & 2 137.70 & 0.41 & 0.44 \\
\textbf{BFGS} & \textbf{CMA} & \textbf{13} & \textbf{10} & -2.60 & 17 152.06 & 11 990.78 & 7 603.76 & 0.56 & 0.30 \\
\textbf{BFGS} & \textbf{CMA} & \textbf{13} & \textbf{20} & -2.40 & 69 235.76 & 42 097.00 & 35 700.52 & 0.48 & 0.39 \\
\textbf{BFGS} & \textbf{CMA} & \textbf{14} & \textbf{2} & -5.20 & 752.24 & 368.06 & 272.58 & 0.64 & 0.51 \\
\textbf{BFGS} & \textbf{CMA} & \textbf{14} & \textbf{3} & -5.40 & 1 285.14 & 600.68 & 475.46 & 0.63 & 0.53 \\
\textbf{BFGS} & \textbf{CMA} & \textbf{14} & \textbf{5} & -5.40 & 2 449.12 & 1 125.06 & 895.84 & 0.63 & 0.54 \\
\textbf{BFGS} & \textbf{CMA} & \textbf{14} & \textbf{10} & -5.40 & 7 147.98 & 3 614.42 & 2 520.06 & 0.65 & 0.49 \\
\textbf{BFGS} & \textbf{CMA} & \textbf{14} & \textbf{20} & -5.40 & 22 830.74 & 12 444.36 & 7 853.46 & 0.66 & 0.45 \\ \hline\hline
\textbf{CMA} & \textbf{BFGS} & \textbf{8} & \textbf{3} & 0.80 & 312.28 & 296.64 & 237.46 & 0.24 & 0.05 \\
CMA & BFGS & 8 & 5 & 0.40 & 7 386.31 & 5 945.52 & 5 965.12 & 0.19 & 0.20 \\
CMA & BFGS & 8 & 10 & 0.60 & 10 671.71 & 7 176.62 & 7 220.73 & 0.32 & 0.33 \\
CMA & BFGS & 9 & 10 & 0.60 & 14 641.85 & 11 104.63 & 11 167.32 & 0.24 & 0.24 \\
CMA & BFGS & 9 & 20 & 0.60 & 33 784.53 & 26 264.40 & 46 188.77 & -0.27 & 0.22 \\\hline
\end{tabular}
\end{center}
\end{table}

We also warm-start the step-size $\sigma$ of CMA-ES from the internal states of BFGS. While the initial value of $\sigma$ can be set directly to its counterpart in BFGS - the length of steps obtained from BFGS' internal line search procedure, this setting usually leads to unsatisfactory performance since BFGS' line search procedure sometimes exhibits unstable behavior, thereby making the initial $\sigma$ of CMA-ES improper. Instead, we consider averaging the step length\footnote{Strictly speaking, for variable-metric optimization methods, e.g., BFGS and CMA-ES, the Mahalanobis metric is the most sensible choice for computing the step length. In our experiment, we choose the Euclidean metric for simplicity, which works quite well according to the experimental result.} of BFGS over the $n$ most recent iterations from the switching point, i.e., $\sigma =n^{-1}\sum_{j=0}^{n-1} \lVert \mathbf{x}_j - \mathbf{x}_{j+1} \rVert$,
where $\mathbf{x}_j$ is the $j$-th recent search point ($\mathbf{x}_0$ is the last point of BFGS before the switching) in the trajectory of BFGS.

We compare the effect of these different parts of the warm-starting procedure on the performance of a dynamic combination of BFGS and CMA-ES. In Fig.~\ref{fig:ERTs-bfgs-cmaes}, we show the resulting ERT-curves on F14 in 2D. From this figure, it is clear to see that switching from BFGS to CMA-ES is beneficial to avoid the numerical issues near the end of the optimization procedure, but the details of the switching procedure have a large effect on the overall performance. We can see that only warm-starting the initial point is not sufficient to beat the CMA-ES itself, as the step size will be much too large (Fig.~\ref{fig:countour_cma_basic_new}, the leftmost sub-plot). Utilizing the warm-starting procedures for both step-size and covariance matrix, we observe a $64\%$ actual performance gain over the \vbss (which is CMA-ES in this case). 

This warm-starting procedure can be applied to all use-cases where the data-driven analysis from Section~\ref{sec:data-driven-analysis} found a combination of BFGS and CMA-ES to be optimal. This corresponds to a total of 23 function-dimension pairs. Note that a similar warm-starting procedure can be applied for the reverse switch (from CMA-ES to BFGS) by passing on the best-so-far and the current covariance matrix. This results in a further 5 use-cases.
The resulting ERT values for these (function, dimension) pairs are summarized in Table~\ref{tab:bfgs_cmaes}. From this table, we see that 25 out of 28 of the dynamic use-cases do indeed improve over the \vbss. 
However, there are three functions where this is not the case. Notably, for function $F8$ in dimension 20, the switch from BFGS to CMA-ES does not perform well, with barely any successful runs. This highlights a disadvantage of the ERT-based calculations to select use-cases: if there is only one successful run of an algorithm, the ERT curve after a particular target can be quite misleading. 

\begin{figure}[t]
    \centering
    \includegraphics[width=.9\textwidth]{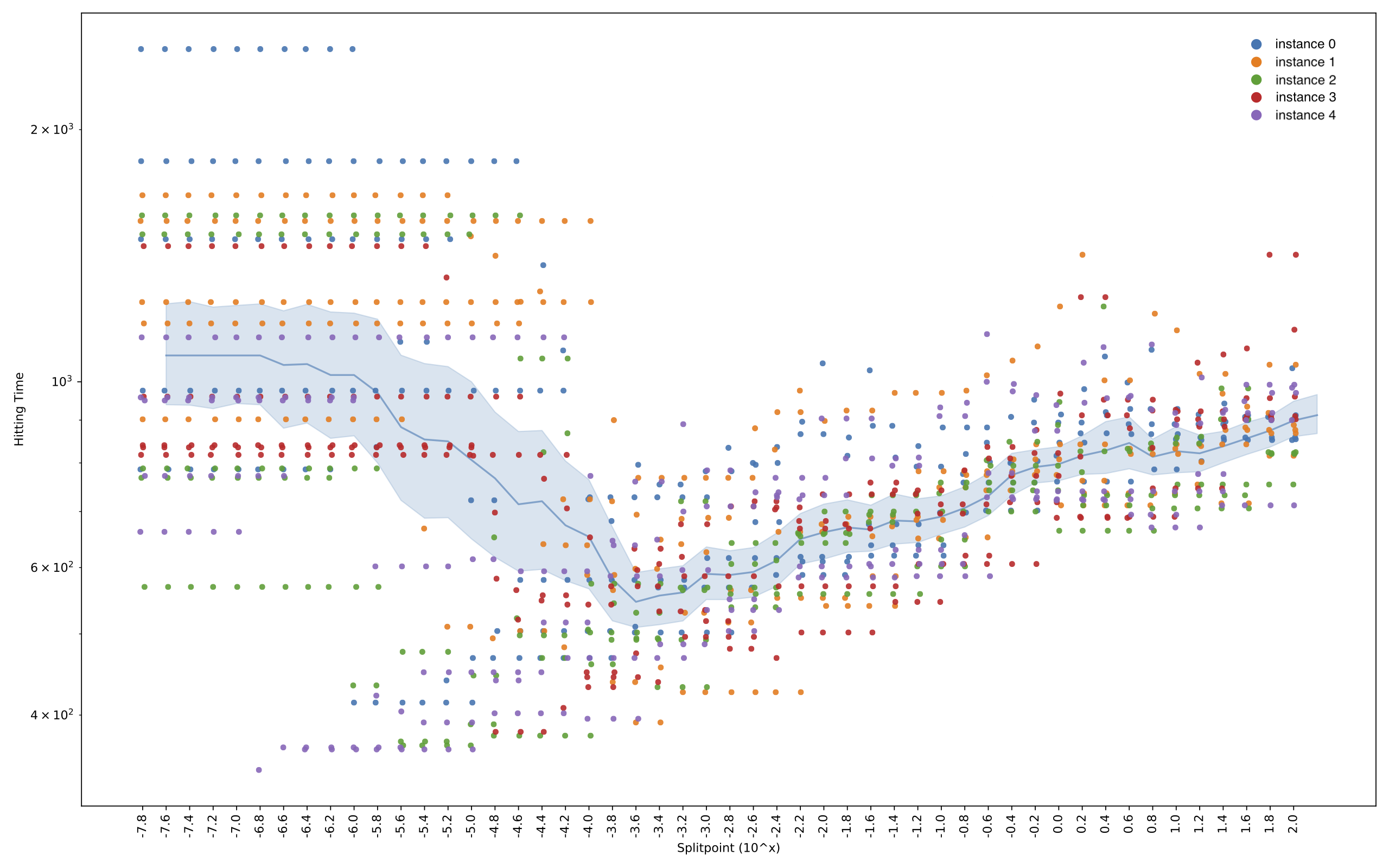}
    \caption{When switching from BFGS to CMA-ES on $F10$ in 5D, the mean and standard deviation of the hitting time are plotted against different values for $\tau$. The hitting times are obtained from five independent runs of this \vbsd on each of the first five instances of $F10$, where the hitting times for each $\tau$ value are depicted as dots with each problem instance being color-coded. 
    \label{fig:F10_5D_split_impact}
    }
\end{figure}

\subsection{Sensitivity Analysis: Impact of the Switching Point}
In the final experiment, we check how the actual performance of a \vbsd is sensitive to the switching point. To conduct this sensitivity analysis, we take the case of switching from BFGS to CMA-ES on $F10$ in 5D, where we observe a significant actual performance gain when using the theoretically derived switching point. We investigate the sensitivity of the switching point $\tau$ by varying its value in a wide range $\{10^i\colon i \in \{-7.8, -7.6, \ldots, 1.8, 2\}\}$. The result is depicted in Fig.~\ref{fig:F10_5D_split_impact}, where it is obvious that the best actual performance of switching from BFGS to CMA-ES was obtained with $\tau = 10^{-3.6}$. Interestingly, when the switch point is smaller than $10^{-4}$, the empirical distribution of the individual hitting times of the \vbsd is substantially different from those with a much larger switch point, showing a more bi-modal distribution, which implies the existence of a transition phase of the behavior of the \vbsd.

This behavior seems to indicate that while switching between algorithms can be beneficial, there is yet more potential to be gained by using run-specific information to determine when to switch between configurations. This could be achieved by using both the internal algorithm state and local algorithm features to create a much more granular selection/configuration policy. Exploratory studies into the potential of these per-run dynamic algorithm selection methods show promising performance~\cite{DBLP:conf/ppsn/KostovskaJVNWED22}.

\section{Conclusions and Future Work}\label{sec:concl}
While we have shown that dynamic algorithm selection can be highly beneficial in certain scenarios, it is clear that much more research is needed to be able to generalize these results to more complex settings. In particular, we have shown that the procedures used to warm-start the algorithms are critical to achieving the potential improvements seen in the performance data of the individual algorithms. However, extending this to more algorithms will require an extendable way to effectively make use of the information collected during the previous phases of the optimization procedure. 

Another major challenge in generalizing the switching between algorithms is the determination of the switch point~\cite{DBLP:conf/cec/RohlerY21}. In this work, we use a fixed-target perspective, based on target precision, which necessarily breaks the black-box assumption present in typical use-cases of evolutionary computation methods. As such, the determination of when to switch between algorithms should be studied in more detail, for example, from the area of dynamic landscape analysis~\cite{JankovicED21}. In this way, the limitation of using a single switch point could also be relaxed, allowing the resulting algorithm to exploit local search more effectively.

Switching between different algorithms can be seen as a special case of the dynamic algorithm configuration problem, where in addition to the choice of the algorithm one also includes the hyperparameters of the algorithms in the design process. This inclusion allows for further specialization of the individual algorithms, thus leading to even more potential differences to exploit in the dynamic context. However, this comes at the cost of increased computational complexity since a purely data-driven approach as used in this work is computationally infeasible as soon as continuous hyperparameters are introduced. Different techniques need to be used for this more general case. At present, existing works focus mainly on reinforcement learning approaches~\cite{BiedenkappBEHL20}.
Switching between different algorithms can be seen as a special case of the dynamic algorithm configuration problem, where in addition to the choice of the algorithm one also includes the hyperparameters of the algorithms in the design process. This inclusion allows for further specialization of the individual algorithms, thus leading to even more potential differences to exploit in the dynamic context. However, this comes at the cost of increased computational complexity since a purely data-driven approach as used in this work is computationally infeasible as soon as continuous hyperparameters are introduced. Different techniques need to be used for this more general case. At present, existing works focus mainly on reinforcement learning approaches~\cite{SharmaKLK19, BiedenkappBEHL20, Biedenkapp0KHD22}.

\section*{Acknowledgements}
This work was supported by a public grant as part of the
Investissements d'avenir project, reference ANR-11-LABX-0056-LMH, LabEx LMH, by the Paris Ile-de-France Region (AlgoSelect project), and via CNRS INS2I project RandSearch. 

\newcommand{\etalchar}[1]{$^{#1}$}

\end{document}